\documentclass{bmvc2k}
\usepackage{adjustbox}
\usepackage{multirow}
\usepackage{multicol}
\usepackage{booktabs}
\usepackage{multirow}
\usepackage{adjustbox}
\usepackage{booktabs}
\usepackage{tabularx}
\usepackage{array}
\usepackage{colortbl}
\usepackage{makecell}
\usepackage{graphicx}
\usepackage{enumitem}
\usepackage{arydshln}
\usepackage{wrapfig,lipsum,booktabs}

\title{Realism to Deception: Investigating
Deepfake Detectors Against Face
Enhancement}

\addauthor{Muhammad Saad Saeed}{msaads@umich.edu}{1}
\addauthor{Ijaz Ul Haq}{ijazhaq@umich.edu}{1}
\addauthor{Khalid Malik}{drmalik@umich.edu}{1}

\addinstitution{
 SMILES Lab,\\
College of Innovation \& Technology,\\
University of Michigan-Flint, Flint, USA
}

\runninghead{SAEED, HAQ, MALIK}{REALISM TO DECEPTION}

\begin{document}

\maketitle

\begin{abstract}
Face enhancement techniques are widely used to enhance facial appearance. However, they can inadvertently distort biometric features, leading to significant decrease in the accuracy of deepfake detectors. This study hypothesizes that these techniques, while improving perceptual quality, can degrade the performance of deepfake detectors. To investigate this, we systematically evaluate whether commonly used face enhancement methods can serve an anti-forensic role by reducing detection accuracy. We use both traditional image processing methods and advanced GAN-based enhancements to evaluate the robustness of deepfake detectors. We provide a comprehensive analysis of the effectiveness of these enhancement techniques, focusing on their impact on Na\"ive, Spatial, and Frequency-based detection methods. Furthermore, we conduct adversarial training experiments to assess whether exposure to face enhancement transformations improves model robustness. Experiments conducted on the FaceForensics++, DeepFakeDetection, and CelebDF-v2 datasets indicate that even basic enhancement filters can significantly reduce detection accuracy achieving ASR up to 64.63\%. In contrast, GAN-based techniques further exploit these vulnerabilities, achieving ASR up to 75.12\%. Our results demonstrate that face enhancement methods can effectively function as anti-forensic tools, emphasizing the need for more resilient and adaptive forensic methods.
\end{abstract}

\section{Introduction}
\label{sec:intro}
Recent advancements in deepfake generation techniques have enabled the creation of synthetic media to the extent that it now appears almost indistinguishable from reality \cite{sha2023deep}. Although these breakthroughs have played a positive role in entertainment, and other applications, they also pose significant threats, including misinformation, fraud, and privacy violations \cite{kwok2021deepfake}. Researchers around the world have developed various deepfake detection methods including na\"ive \cite{tan2019efficientnet, rossler2019faceforensics++}, spatial \cite{ni2022core, yan2023ucf}, and frequency \cite{qian2020thinking, luo2021generalizing}  based methods. Despite their promising performance, deepfake detectors remain vulnerable to adversarial attacks that exploit their sensitivity to carefully crafted perturbations. These distortions are often imperceptible to human eye, but can significantly reduce detection accuracy \cite{uddin2023robust}.

\begin{figure}[t]
\centering
\includegraphics [width=4.5 in]{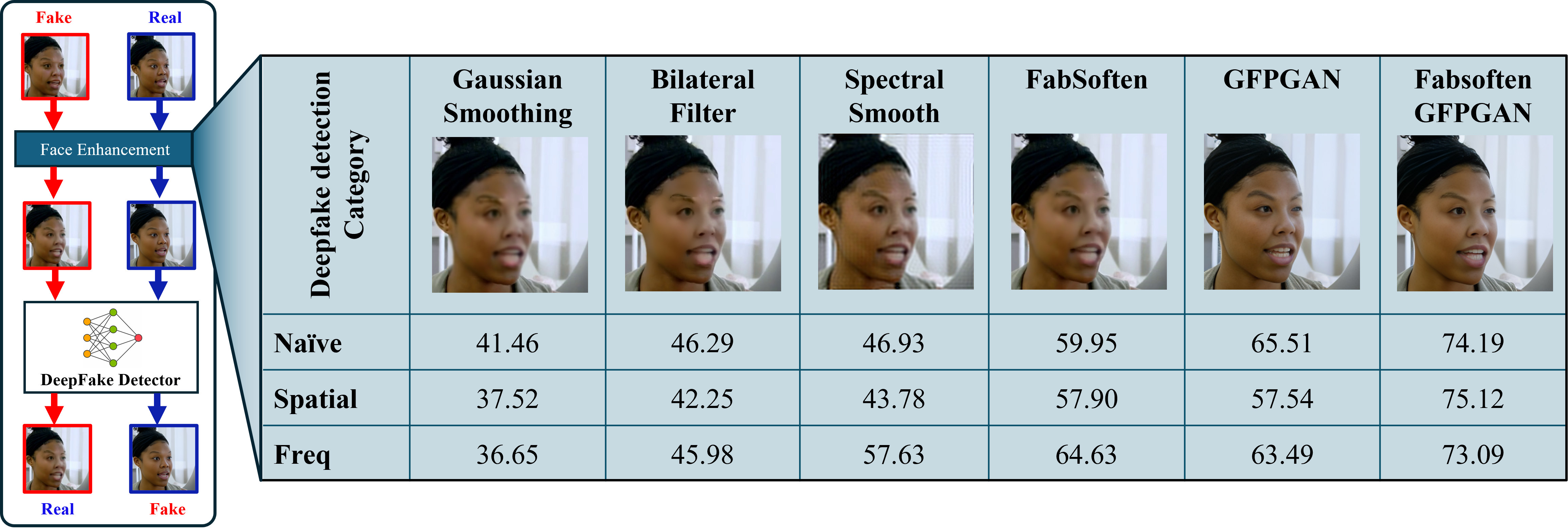}
\caption{ Using face enhancement as anti-forensic in deepfake detectors (left).Naïve, Spatial, and Frequency-based detection methods with average attack success rates (ASR$^{\uparrow}$) across various enhancement methods (right). Best viewed in color and zoomed in.}
\label{fig:fig_1}
\end{figure}

Unlike the existing studies that explore adversarial attacks specifically designed for deepfake detection models, we focus on methods from face enhancement domain such as skin smoothing, and detail enhancement. These face enhancement techniques, while designed to improve perceptual quality, can act as anti-forensic attack by concealing artifacts and cues that deepfake detectors rely upon. This dual role not only challenges the robustness of deepfake detection systems but also introduces an additional layer of complexity by misleading human perception. Figure \ref{fig:fig_1}  shows how face enhancement reduces accuracy across various detectors, raising concerns about their anti-forensic potential and implications for both automated systems and human judgment. To further explore this dual characteristics of face enhancement methods, this paper answers the following research questions:

\begin{itemize}[leftmargin=*]
\item Are face enhancement filters effective as anti-forensic attacks?
\item To what extent do face enhancement methods exhibit anti-forensic capabilities against Naïve, Spatial, and Frequency-based deepfake detectors?
\item What are the trade-offs between perceptual quality and detection evasion when applying face enhancement methods?
\item Can adversarial training mitigate the anti-forensic effects of face enhancement methods?
\end{itemize}

By investigating these questions, we aim to bridge the gap between forensic frameworks and real-world scenarios where face-enhancement methods may degrade detection performance, providing a deeper understanding of deepfake detection vulnerabilities. Our contributions are summarized below:

\begin{enumerate}[leftmargin=*]
  \item We analyze and demonstrate that face enhancement techniques can effectively deceive state-of-the-art deepfake detectors.
  \item We further analyze that applying face enhancement filters not only evades detection but also enhances the visual realism of deepfake videos by minimizing artifact traces, making them imperceptible to the human eye.
  \item We experimentally prove that these enhancement techniques can generalize across diverse categories of detectors, including Na\"ive, Spatial, and Frequency-based methods, highlighting their anti-forensic impact on detection performance.
\end{enumerate}

\section{Literature review }
Several studies have explored adversarial attacks within the broader context of computer vision and neural networks \cite{uddin2023deep}. An early study by Carlini and Wagner \cite{carlini2017towards} highlighted the vulnerability of neural networks to imperceptible perturbations. Similarly, Li et al. \cite{li2023frequency} proposed frequency domain regularization for iterative adversarial attacks, crafting imperceptible yet highly effective perturbations. Beyond these foundational studies, several studies focus specifically on adversarial challenges within the domain of deepfake detection \cite{dong2023restricted, gowrisankar2024adversarial}. For instance, Hussain  et al. \cite{hussain2021adversarial}  propose adversarial perturbations that can bypass DNN-based Deepfake detectors in both white- and black-box attack scenarios, even under compression, posing a significant real-world threat. Ain  et al. \cite{ain2024exposing} demonstrate that visually natural perturbations, such as perceptual facial moles, can significantly undermine detection accuracy. Hou  et al. \cite{hou2023evading} propose statistical consistency attack named StatAttack, that minimizes the statistical differences between real and fake images using natural degradations and distribution-aware loss effectively bypassing spatial and frequency-based deepfake detectors. Carlini et al.~\cite{carlini2020evading} explored latent-space perturbations within the generative model to produce adversarial images that evade deepfake detectors. Huang et al.~\cite{huang2020fakepolisher} proposed FakePolisher, a learning-based reconstruction method that suppresses GAN synthesis footprints, significantly degrading state-of-the-art deepfake detectors and revealing their reliance on low-level artifacts. 
Ivanovska et al.~\cite{ivanovska2024vulnerability} examined single-image deepfake detector vulnerabilities to black-box attacks via Denoising Diffusion Models, showing that even one guided denoising step can markedly lower detection rates without visible changes. Despite these advances, most existing studies focus predominantly on mathematical perturbations, such as noise-based or gradient-guided methods \cite{meng2024ava}. They often neglect the potential of practical, real-world techniques like face enhancement filters, which can serve as both perceptual improvement tools and detection evading mechanisms. This gap highlights the need for further research into practical attacks that can effectively bypass deepfake detection systems while maintaining high perceptual realism.

\section{Design of the empirical study}

\subsection{Datasets and evaluation metrics }
We evaluate face enhancement methods against deepfake detectors using FaceForensics++ (c23), DeepFakeDetection (c23), and CelebDF-v2. Every fifth frame from real and fake videos is sampled for balanced representation, following standard train-test splits. Perceptual quality is measured with SSIM, PSNR, and LPIPS, while detection degradation is reported via attack success rate (ASR), defined as the proportion of fake samples misclassified as real.

\subsection{Face enhancement methods}
To conduct this empirical study, we select six face enhancement techniques and apply them to regenerate the deepfake datasets accordingly. These techniques span classical spatial smoothing, transform-domain smoothing, and generative restoration approaches.\\
\noindent
\textbf{\textit{M1: Gaussian Smoothing}} \cite{taubin1995curve} is a common face enhancement technique used to reduce imperfections such as blemishes, wrinkles, and uneven textures by applying a weighted average to surrounding pixels. 


\vspace{0.3em}
\noindent
\textbf{\textit{M2: Bilateral Filtering}} \cite{tomasi1998bilateral} is another smoothing technique that reduces the prominence of uneven skin tones while preserving important edge details on the face. Bilateral filtering operates by averaging pixel values based on both spatial proximity and intensity similarity.

\vspace{0.3em}
\noindent
\textbf{\textit{M3: Spectral Smoothing}} \cite{ahmed2006discrete}  is a frequency-domain low-pass filtering technique that selectively suppresses high-frequency components. The image is transformed via the Discrete Fourier Transform (DFT).

\vspace{0.3em}
\noindent
\textbf{\textit{M4: FabSoften}} \cite{velusamy2020fabsoften} enhances facial aesthetics by dynamically smoothing skin blemishes, guided by facial attributes such as blemish density and texture coarseness, and restores natural skin texture through wavelet-based manipulations. 

\vspace{0.3em}
\noindent
\textbf{\textit{M5: GFPGAN}} \cite{wang2021towards} is a blind face restoration approach leveraging a Generative Facial Prior from a pretrained StyleGAN2 model. GFPGAN jointly performs degradation removal and detail restoration in a single forward pass. 

\vspace{0.3em}
\noindent
\textbf{\textit{M6: FabSoften + GFPGAN}} combines wavelet-domain smoothing with generative restoration. First, we apply FabSoften to suppress high-frequency blemishes, followed by GFPGAN to reconstruct perceptually convincing facial details.

\subsection{Deepfake detectors}
For this empirical study, we selected six deepfake detection methods, comprising two prominent deepfake detectors of each category—Na\"ive, Spatial, and Frequency-based. 

\subsubsection{Na\"ive deepfake detectors}
These methods generally use CNNs to directly differentiate between deepfake and genuine media. While computationally efficient, they are vulnerable to reductions in accuracy caused by subtle modifications and enhancements that exhibit anti-forensic effects.\\
\noindent
\textbf{ \textit{EfficientNet-B4\cite{tan2019efficientnet}}} is one of the widely used CNN in deepfake detection due to its highly optimized architecture, which balances model depth, width, and resolution for superior performance. Its compound scaling approach allows it to efficiently capture fine-grained spatial details and subtle manipulations in deepfake images and videos.\\
\noindent
\textbf{\textit{Xception \cite{rossler2019faceforensics++}}}
is popular in deepfake detection due to its superior feature extraction capabilities. Xception leverages depth-wise separable convolutions to efficiently capture spatial and channel-wise relationships, making it highly effective in identifying deepfake artifacts.

\subsubsection{Spatial deepfake detectors}
These detectors leverage deep neural networks to analyze pixel-level information, focusing on spatial inconsistencies in texture, or lighting making them effective for high-resolution images.\\
\noindent
\textbf{\textit{CORE \cite{ni2022core}}} captures different representations of the same sample via different augmentations. The framework explicitly enforces representation consistency across different augmented views through a cosine-based Consistency Loss.\\
\noindent
\textbf{\textit{UCF \cite{yan2023ucf}}} employs a multi-task disentanglement framework to tackle two challenges: overfitting to irrelevant features and overfitting to method-specific textures. By extracting shared features, the framework addresses these issues while improving generalization capability. 

\subsubsection{Frequency deepfake detectors} Frequency-based methods target manipulations by identifying anomalies in low or high frequency components. They are particularly robust against compressed inputs, but may struggle when modifications involve frequency-domain alterations.
\noindent
\textbf{\textit{F$^{3}$Net \cite{qian2020thinking}}} employs a cross-attention two-stream network to collaboratively learn frequency-aware features from two branches: FAD and LFS. The FAD module divides the input image using learnable frequency bands to extract frequency-aware components. The LFS module captures localized frequency statistics to highlight differences between real and fake faces.\\
\noindent
\textbf{\textit{SPSL \cite{luo2021generalizing}}} integrates spatial image features with the phase spectrum to detect up-sampling artifacts in face forgeries, enhancing transferability and generalization for face forgery detection. This method employs Xception as the backbone architecture.

\begin{table}[t]
    \centering
    \caption{ASR of face enhancement methods against deepfake detectors. Results are reported across FaceForensics++ (FF++), DeepfakeDetection (DFD), and Celeb-DFv2 (CDFv2).}
    \begin{adjustbox}{width=\textwidth}
    \begin{tabular}{ccccccccccccccccccccccc}
        \toprule
        \multirow{3}{*}{\textbf{Category}} & \multirow{3}{*}{\textbf{Method}} 
        & \multicolumn{3}{c}{\textbf{Gaussian Blur}}
        & \multicolumn{3}{c}{\textbf{Bilateral Filter}}
        & \multicolumn{3}{c}{\textbf{Spectral Smooth}}
        & \multicolumn{3}{c}{\textbf{FabSoften}}
        & \multicolumn{3}{c}{\textbf{GFP-GAN}}
        & \multicolumn{3}{c}{\textbf{FabSoften+GFPGAN}} \\
        \cmidrule(lr){3-5} \cmidrule(lr){6-8} \cmidrule(lr){9-11} \cmidrule(lr){12-14} \cmidrule(lr){15-17} \cmidrule(lr){18-20}
        & & \textbf{FF++} & \textbf{DFD} & \textbf{CDFv2} 
          & \textbf{FF++} & \textbf{DFD} & \textbf{CDFv2} 
          & \textbf{FF++} & \textbf{DFD} & \textbf{CDFv2} 
          & \textbf{FF++} & \textbf{DFD} & \textbf{CDFv2} 
          & \textbf{FF++} & \textbf{DFD} & \textbf{CDFv2} 
          & \textbf{FF++} & \textbf{DFD} & \textbf{CDFv2} \\
        \midrule
        \multirow{2}{*}{Na{\"i}ve} 
        & EffNetB4 
        & 40.78 & 42.85 & 51.36 
        & 47.91 & 48.97 & 58.17
        & 46.64 & 39.92 & 41.46
        & 56.77 & 64.46 & 62.53
        & 68.85 & 55.49 & 54.43
        & 72.12 & 87.39 & 78.60 \\
        & Xception 
        & 42.15 & 41.88 & 52.68 
        & 44.66 & 44.15 & 61.37
        & 47.22 & 34.89 & 46.07
        & 63.12 & 65.65 & 64.08
        & 62.16 & 58.21 & 51.70
        & 76.25 & 85.97 & 77.29 \\
        \hline
        \multirow{2}{*}{Spatial} 
        & Core 
        & 38.99 & 44.44 & 49.10 
        & 43.31 & 50.86 & 53.67
        & 42.48 & 40.17 & 41.37
        & 55.83 & 63.96 & 61.19
        & 59.12 & 54.73 & 56.46
        & 76.95 & 80.33 & 79.22 \\
        & UCF 
        & 36.05 & 43.82 & 48.88 
        & 41.18 & 50.26 & 52.13
        & 45.07 & 44.14 & 43.68
        & 59.97 & 62.03 & 64.63
        & 55.96 & 52.65 & 54.87
        & 73.29 & 80.50 & 78.97 \\
        \hline
        \multirow{2}{*}{Frequency} 
        & F$^{3}$Net 
        & 36.71 & 42.57 & 50.02 
        & 43.91 & 49.81 & 55.87
        & 56.30 & 41.78 & 46.37
        & 63.81 & 63.39 & 69.92
        & 62.80 & 53.54 & 57.25
        & 71.34 & 80.39 & 80.16 \\
        & SPSL 
        & 36.59 & 41.34 & 48.66 
        & 48.05 & 55.55 & 50.92
        & 58.96 & 48.81 & 51.87
        & 65.45 & 65.35 & 68.38
        & 64.17 & 55.97 & 60.82
        & 74.83 & 80.33 & 83.75 \\
        \bottomrule
    \end{tabular}
    \end{adjustbox}
    \label{tab:asr}
\end{table}

\begin{figure}[t]
    \centering

    \includegraphics[width=0.6\textwidth]{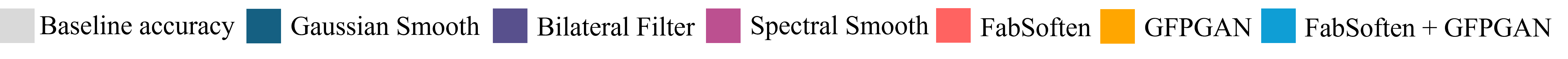}

    \begin{minipage}{0.32\textwidth}
        \centering
        \includegraphics[width=\textwidth]{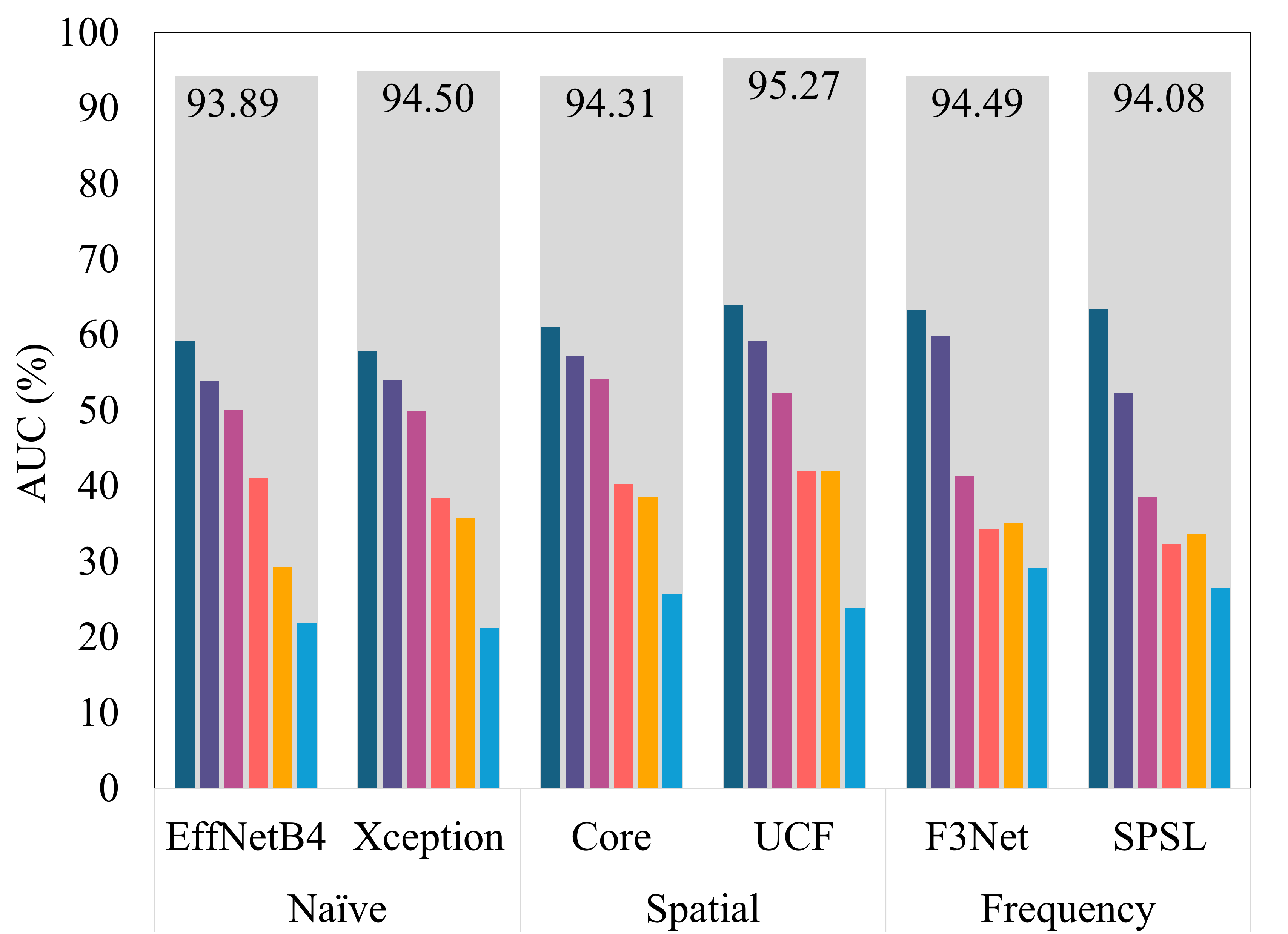}

        {\small (a) FF++}
    \end{minipage}
    \hfill
    \begin{minipage}{0.32\textwidth}
        \centering
        \includegraphics[width=\textwidth]{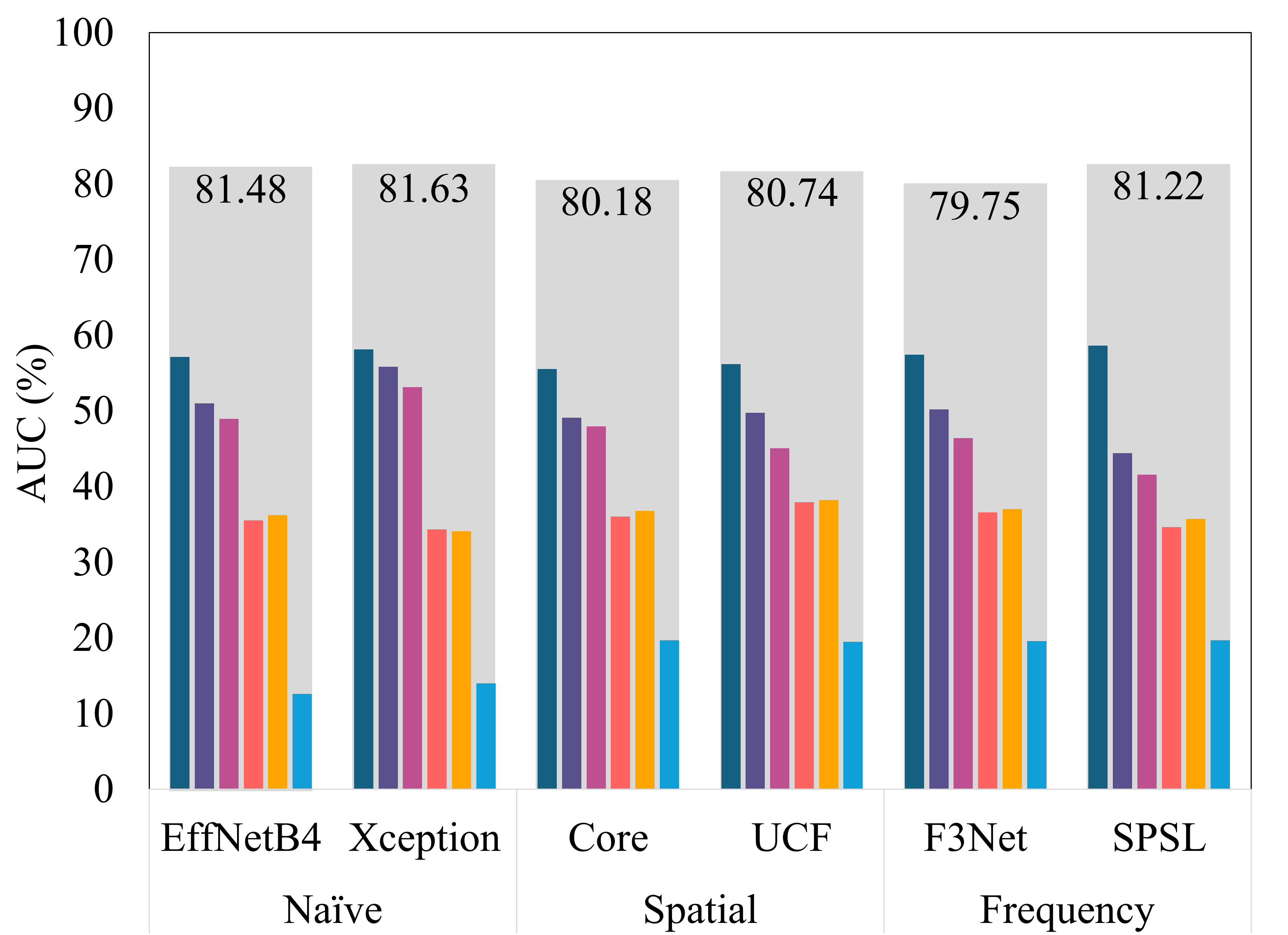}

        {\small (b) DFD}
    \end{minipage}
    \hfill
    \begin{minipage}{0.32\textwidth}
        \centering
        \includegraphics[width=\textwidth]{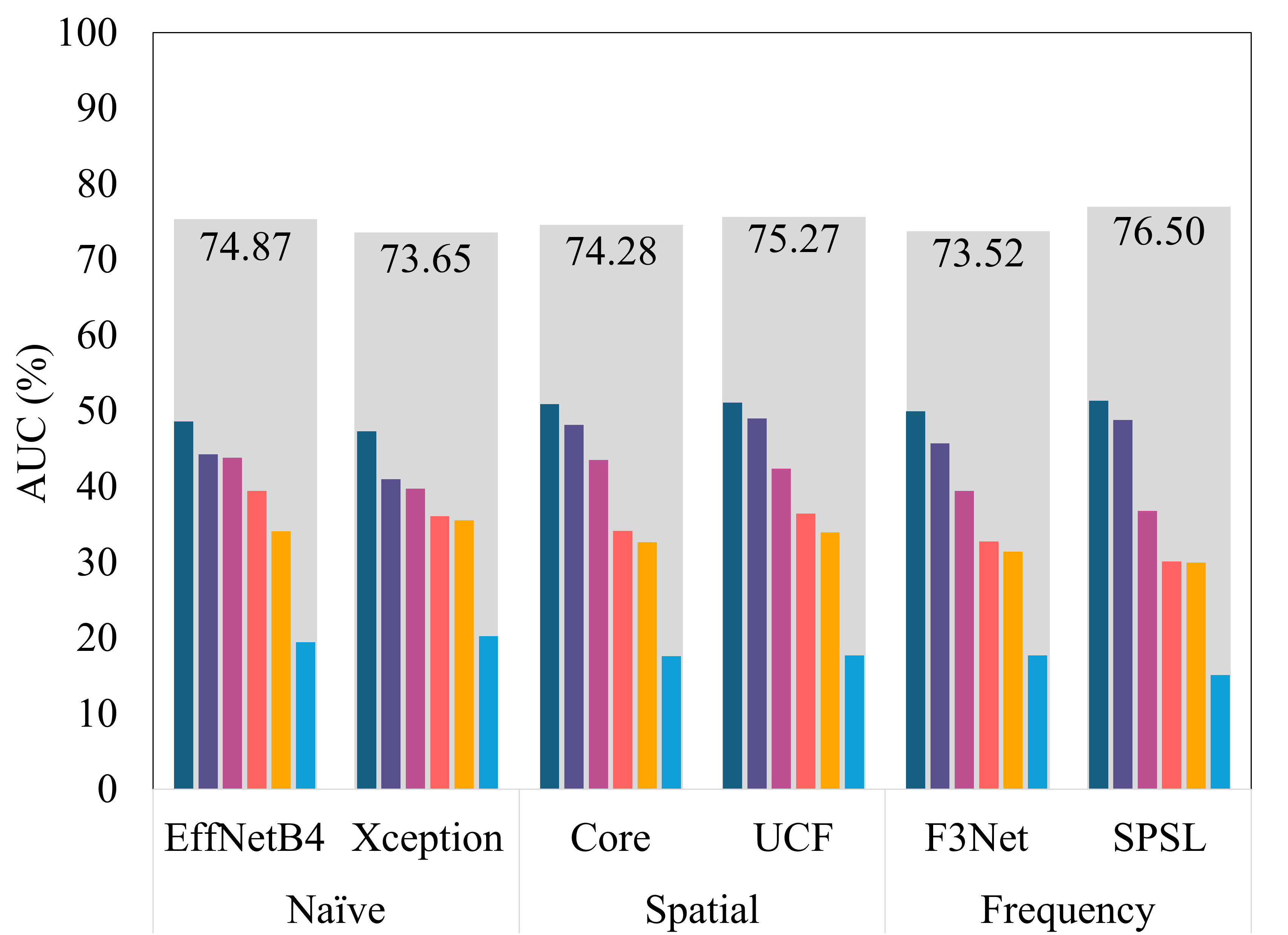}

        {\small (c) CelebDF-v2}
    \end{minipage}

    \caption{Deepfake detector AUC across FaceForensics++ (FF++), DeepfakeDetection (DFD), and Celeb-DFv2 datasets. Best viewed in color and zoomed in.}
    \label{fig:accuracy_bar}
\end{figure}

\section{Experimental evaluations}

\subsection{Deepfake detection performance under face enhancement}
To evaluate the effectiveness of face enhancement techniques as anti-forensic attacks, we compute the ASR across three categories of deepfake detectors: Naïve, Spatial, and Frequency-based. As shown in Table~\ref{tab:asr}, FabSoften+GFPGAN achieves the highest ASR across all categories, indicating its superior capability in bypassing detection systems. Specifically, it achieves an average ASR of 74.19\%, 75.12\%, and 73.09\% against Naïve, Spatial, and Frequency-based detectors, respectively. In contrast, Gaussian smoothing and Bilateral filtering exhibit the lowest ASRs, with Gaussian smoothing averaging only 41.47\%, 37.52\%, and 36.65\% across the same detector categories. The standalone FabSoften method shows moderate effectiveness, consistently outperforming classical smoothing but falling short of GAN-based enhancement. Figure \ref{fig:accuracy_bar} reports AUC (\%) for each method across detectors. The grey bar indicates the baseline AUC for reference. A consistent trend of reduced resilience to enhancement methods is observed across all detector categories. These results reflect the vulnerabilities of each detector type: Naïve detectors rely on low-level artifacts disrupted by basic enhancements; Spatial detectors focus on pixel irregularities affected by GAN-based smoothing; Frequency-based detectors detect spectral anomalies, disrupted due to attenuation of high-frequency components in transform domains. 

\begin{figure}[t]
\centering
\includegraphics [width=4.8 in]{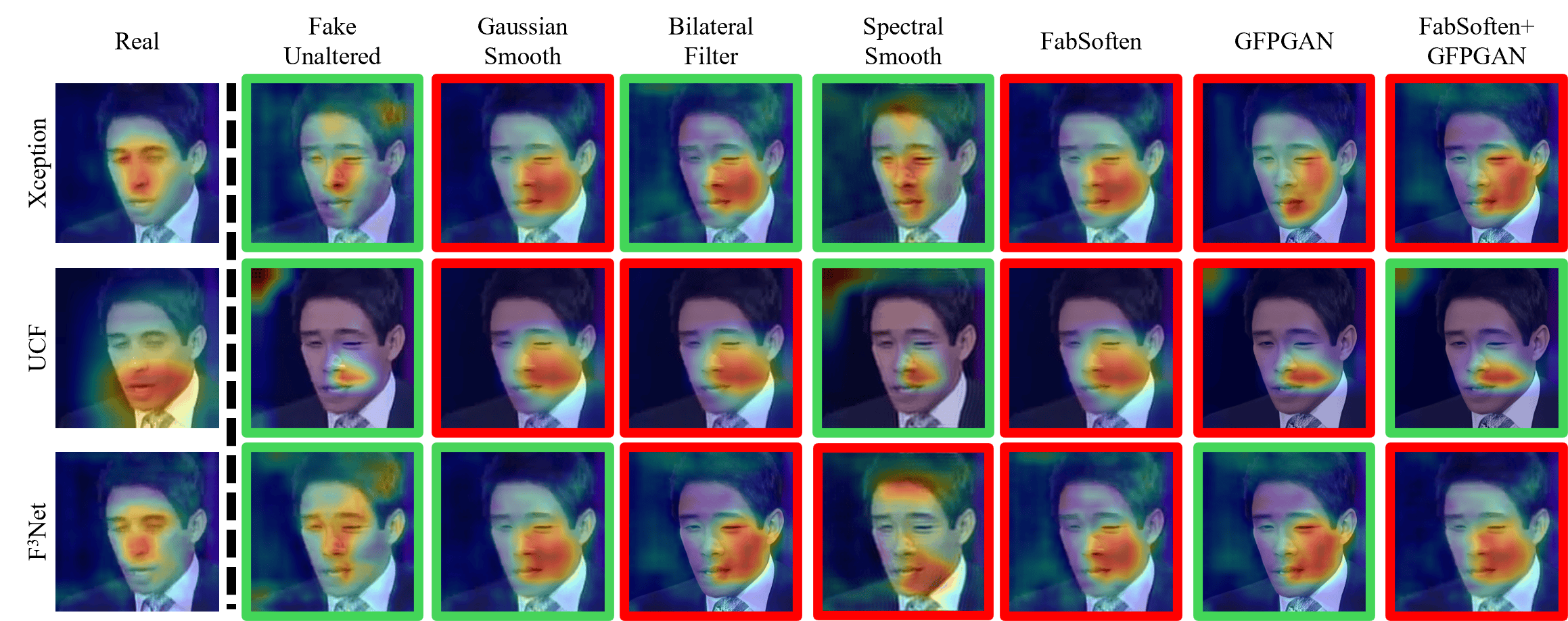}
\caption{Grad-CAM visualization of detector attention. Ground truth: Deepfake. \textbf{Green}: Correct detection (classified as fake), \textbf{Red}: Incorrect detection (misclassified as real).}
\label{fig:gradcam}
\end{figure}


To better understand the internal behavior of the detection models, we analyze Grad-CAM visualizations in Figure~\ref{fig:gradcam} for three representative detectors: Xception, UCF, and F$^3$Net. The heatmaps show that fake samples incorrectly predicted as real often exhibit activation patterns either more similar to genuine samples or deviate from the typical patterns associated with fake images. This trend is especially pronounced in Xception and F$^3$Net, where enhancement techniques appear to shift the models' attention away from fake regions. This shift suggests that face enhancement effectively suppresses or modifies visual cues that detectors rely on, thereby misleading the models into incorrect classifications.

\subsection{Quality assessment of the face enhancement samples}
Generating enhanced deepfake samples through various image processing techniques results in noticeable alterations to the images, as demonstrated by the quantitative assessment in Table~\ref{tab:metrics_tables}. Techniques like Bilateral Filtering and Gaussian Smoothing strongly preserve structural and perceptual similarity, with minimal alterations to the original deepfakes. In contrast, Spectral Smoothing, FabSoften, GFPGAN, and FabSoften+GFPGAN make more pronounced changes, yielding the lowest perceptual scores.

\begin{figure}[t]
\centering
\includegraphics [width=4.5 in]{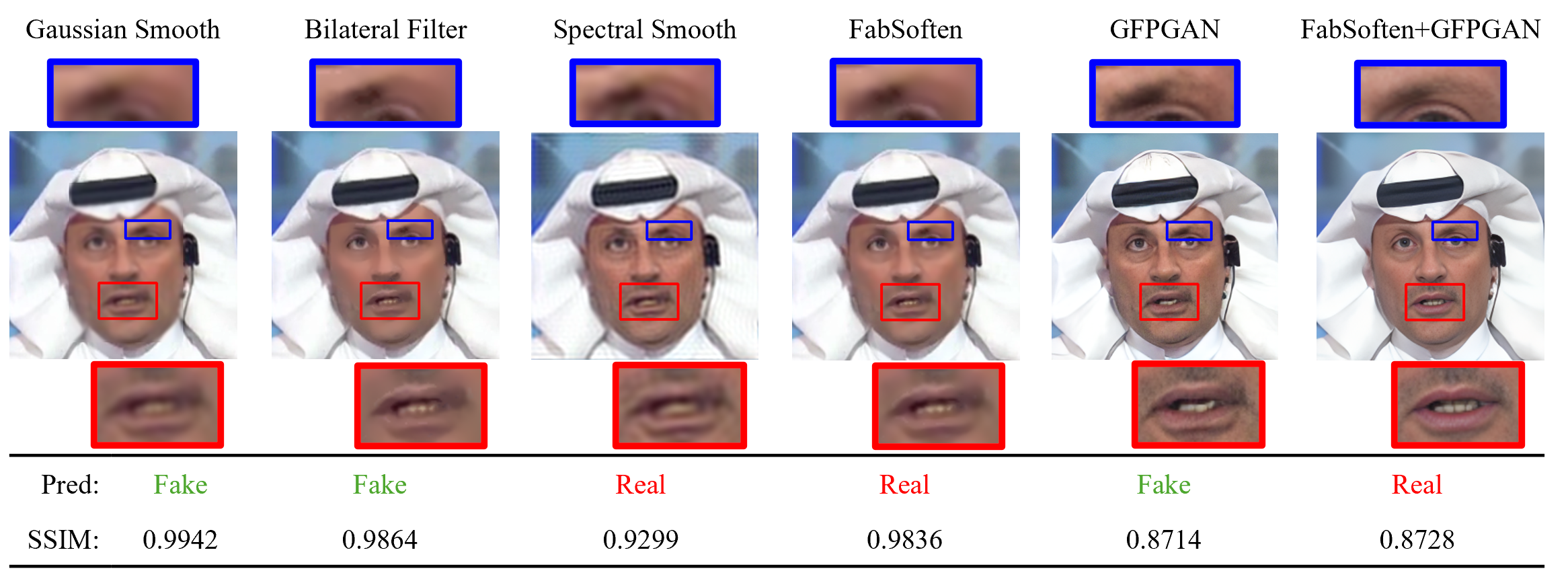}
\caption{Qualitative comparison of face enhancement. Bottom: impact on prediction shift (Fake→Real) and structural preservation (SSIM$^\uparrow$). Best viewed in color and zoomed in.}
\label{percept_fig}
\end{figure}

In Figure \ref{percept_fig}, the impact of different face enhancement methods on facial features, particularly the eyebrows and lips, is clearly visible. It is evident that the Gaussian Smoothing and Bilateral Filtering slightly enhance the facial features while maintaining their structural integrity. The eyebrows and lips appear slightly refined and natural, aligning with the high SSIM (0.9942, 0.9864) and low LPIPS (0.0120, 0.0061) values. On the contrary, Figure~\ref{fig:trade_severe}a illustrate GAN-based enhancements, such as GFPGAN and FabSoften+GFPGAN, that have drastically altered deepfakes by modifying soft biometric traits such as age, skin texture, and facial symmetry. These changes improve the visual quality but also disrupt key forensic cues, causing detectors to misclassify.

\begin{table}[t]
    \centering
    \caption{Quantitative evaluation of generated attacks using perceptual metrics: SSIM$^\uparrow$, PSNR$^\uparrow$ (higher is better), and LPIPS$^\downarrow$ (lower is better).}
    \begin{adjustbox}{width=\textwidth}
    \begin{tabular}{cccccccccccccccccccccccc}
        \toprule
        \multirow{3}{*}{\textbf{Category}} 
        & \multicolumn{7}{c}{\textbf{SSIM$^{\uparrow}$}}
        & \multicolumn{7}{c}{\textbf{PSNR$^{\uparrow}$}}
        & \multicolumn{7}{c}{\textbf{LPIPS$^{\downarrow}$}} \\
        \cmidrule(lr){2-8} \cmidrule(lr){9-15} \cmidrule(lr){16-22}
         & \textbf{FF-DF} & \textbf{FF-F2F} & \textbf{FF-FS} 
         & \textbf{FF-NT} & \textbf{Avg.} & \textbf{DFD} & \textbf{CDFv2} 
         & \textbf{FF-DF} & \textbf{FF-F2F} & \textbf{FF-FS} 
         & \textbf{FF-NT} & \textbf{Avg.} & \textbf{DFD} & \textbf{CDFv2}
         & \textbf{FF-DF} & \textbf{FF-F2F} & \textbf{FF-FS} 
         & \textbf{FF-NT} & \textbf{Avg.} & \textbf{DFD} & \textbf{CDFv2} \\
        \midrule
        \textbf{\textit{M1}} 
        & 0.9847 & 0.9838 & 0.9836 & 0.9831 & 0.9838 & 0.9841 & 0.9914
        & 41.65 & 41.43 & 41.35 & 41.56 & 41.49 & 41.88 & 44.34
        & 0.0380 & 0.0416 & 0.0442 & 0.0389 & 0.0406 & 0.0452 & 0.0212 \\
        \textbf{\textit{M2}} 
        & 0.9951 & 0.9949 & 0.9942 & 0.9938 & 0.9945 & 0.9954 & 0.9945
        & 47.62 & 47.29 & 47.25 & 47.47 & 47.40 & 48.75 & 47.98
        & 0.0061 & 0.0063 & 0.0061 & 0.0058 & 0.0061 & 0.0065 & 0.0087 \\
        \textbf{\textit{M3}} 
        & 0.9355 & 0.9349 & 0.9366 & 0.9395 & 0.9366 & 0.9356 & 0.9454
        & 37.04 & 36.97 & 36.90 & 37.14 & 37.01 & 35.99 & 39.95
        & 0.0924 & 0.0956 & 0.1101 & 0.1095 & 0.1019 & 0.1072 & 0.0550 \\
        \textbf{\textit{M4}} 
        & 0.9514 & 0.9536 & 0.9576 & 0.9511 & 0.9534 & 0.9536 & 0.9597
        & 37.24 & 37.43 & 37.66 & 37.14 & 37.36 & 37.46 & 37.88
        & 0.0575 & 0.0511 & 0.0594 & 0.0561 & 0.0555 & 0.0591 & 0.0599 \\
        \textbf{\textit{M5}} 
        & 0.8734 & 0.8749 & 0.8756 & 0.8745 & 0.8746 & 0.8744 & 0.8605
        & 33.13 & 33.19 & 33.14 & 33.18 & 33.16 & 33.59 & 33.28
        & 0.1271 & 0.1215 & 0.1384 & 0.1264 & 0.1283 & 0.1173 & 0.1493 \\
        \textbf{\textit{M6}} 
        & 0.8656 & 0.8649 & 0.8667 & 0.8666 & 0.8659 & 0.8557 & 0.8654
        & 36.07 & 36.03 & 36.08 & 36.08 & 36.06 & 31.37 & 31.36
        & 0.1172 & 0.1118 & 0.1171 & 0.1149 & 0.1152 & 0.1099 & 0.1008 \\
        \bottomrule
    \end{tabular}
    \end{adjustbox}
    \label{tab:metrics_tables}
\end{table}

\setlength\dashlinedash{1pt}  
\setlength\dashlinegap{1pt}   
\setlength\arrayrulewidth{0.8pt} 

\begin{table}[t]
    \centering
    \caption{Attack success rates versus perceptual quality across various parameter configurations. Bold marks optimal trade-offs.}
    \begin{adjustbox}{width=\textwidth}
    \begin{tabular}{ccc:ccccc:ccc:cccc:ccc}
        \toprule
        \multicolumn{3}{c}{\textbf{Gaussian Smooth}}
        & \multicolumn{5}{c}{\textbf{Bilateral Filter}}
        & \multicolumn{3}{c}{\textbf{Spectral Smooth}}
        & \multicolumn{4}{c}{\textbf{FabSoften}}
        & \multicolumn{3}{c}{\textbf{Combination}}
        \\
        \cmidrule(lr){1-3}
        \cmidrule(lr){4-8}
        \cmidrule(lr){9-11}
        \cmidrule(lr){12-15}
        \cmidrule(lr){16-18}
        \textit{G} & ASR(\%) & SSIM 
        & $\sigma_{color}$ & $\sigma_{space}$ & \textit{d} 
        & ASR(\%) & SSIM 
        & \textit{r}  & ASR(\%) & SSIM 
        & $\alpha_r$ & $\alpha_\epsilon$ & ASR(\%) & SSIM 
        & Methods & ASR(\%) & SSIM \\

        \cmidrule(lr){1-3}
        \cmidrule(lr){4-8}
        \cmidrule(lr){9-11}
        \cmidrule(lr){12-15}
        \cmidrule(lr){16-18}

        3$\times$3 & 36.175 & 0.9898 
        & 30 & 50 & 9 & 42.15 & 0.9971
        & 10 & 61.37 & 0.7219
        & 1 & 5 & 39.73 & 0.9652 & M1 + M5 & 73.35 & 0.8790\\

        5$\times$5 & 36.17 & 0.9850
        & \textbf{60} & \textbf{100} & \textbf{17} & \textbf{44.66} & \textbf{0.9945}
        & 20 & 53.75 & 0.8840
        & 5 & 10 & 56.42 & 0.9518 & M1 + M5 & 71.82 & 0.8725\\

        \textbf{7$\times$7} & \textbf{42.15} & \textbf{0.9838}
        & 80 & 120 & 16 & 58.70 & 0.9447
        & \textbf{40} & \textbf{47.22} & \textbf{0.9366}
        & \textbf{10} & \textbf{5} & \textbf{63.12} & \textbf{0.9534} & M3 + M5 & 74.79 & 0.8553\\

        9$\times$9 & 49.26 & 0.9627
        & 100 & 150 & 17 & 57.25 & 0.9246
        & 60 & 35.47 & 0.9876
        & 10 & 10 & 65.78 & 0.9213 & \textbf{M4 + M5} & \textbf{76.25} & \textbf{0.8659} \\
        \bottomrule
    \end{tabular}
    \end{adjustbox}
    \label{tab:parameter}
\end{table}

\begin{figure}[t]
\centering
\includegraphics[width=4.5 in]{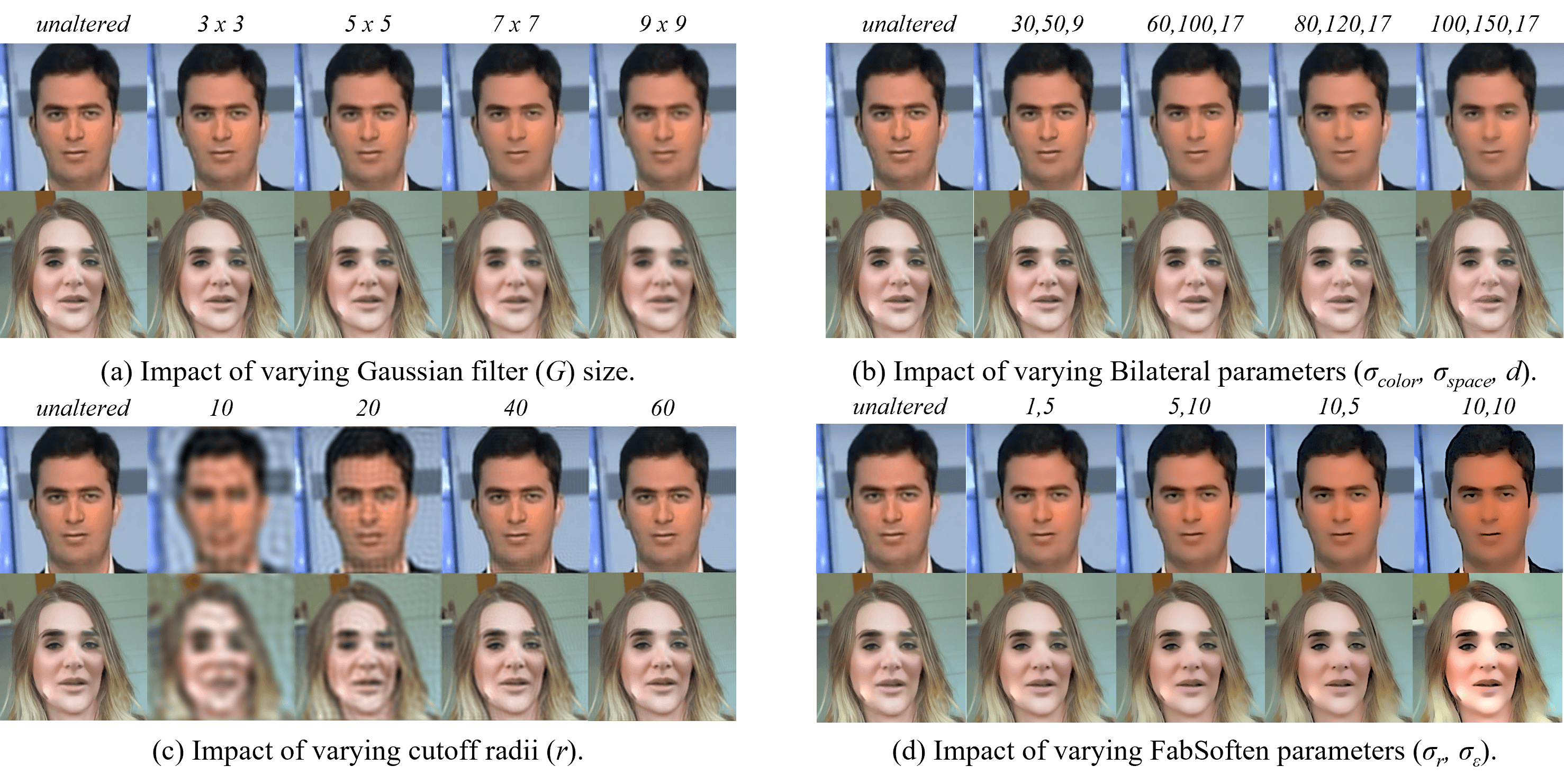}
\caption{Qualitative effects of different parameters on the visual output of face enhancement methods. Best viewed in color.}
\label{fig:ablation}
\end{figure}

\subsection{Ablation study on parameter settings}

This section analyzes different parameter settings and combinations of face enhancement techniques on deepfake detection task, using the Xception model on the FF++ dataset.

\noindent
\textbf {M1:} 
We applied Gaussian smoothing with kernel sizes 3$\times$3 to 9$\times$9 and $\sigma=1.5$. The 7$\times$7 kernel achieved the best trade-off, with ASR = 42.15\% while maintaining high similarity (SSIM = 0.9838). The 9$\times$9 kernel reached the highest ASR (49.26\%) but noticeably degraded quality (SSIM = 0.9627). The visual impact is shown in Figure~\ref{fig:ablation}a.

\noindent
\textbf{M2: }For Bilateral Filtering, we experimented with different values of diameter ($d$), sigma color ($\sigma_{\text{color}}$), and sigma space ($\sigma_{\text{space}}$). As shown in Table~\ref{tab:parameter}, the configuration $d=17$, $\sigma_{\text{color}}=60$, $\sigma_{\text{space}}=100$ achieved the best balance between detection evasion (ASR = 53.99\%) and perceptual similarity (SSIM = 0.9945). Figure~\ref{fig:ablation}b shows the visual impact.

\noindent
\textbf{M3: }We tested spectral smoothing with different cutoff radii (\textit{r}) as given in Table \ref{tab:parameter}. A smaller radii caused excessive blurring and higher ASR, while larger values preserved more detailed but reduced ASR. We selected a radius of 40 as it offered a good balance between concealment (ASR = 42.55\%) and perceptual quality (SSIM = 0.9475). The visual impact of varying radius is shown in Figure~\ref{fig:ablation}c.



\noindent
\textbf{M4:} We evaluated combinations of $\alpha_r$ and $\alpha_\epsilon$ as given in Table~\ref{tab:parameter}. Increasing these parameters progressively improved ASR, reaching a maximum of 65.78\% when both parameters were set to 10, but at the cost of lower SSIM (0.9213). The setting $\alpha_r=10$, $\alpha_\epsilon=5$ provides a favorable trade-off between ASR (63.12\%) and structural similarity (SSIM=0.9534). The visual impact of varying parameters is shown in Figure~\ref{fig:ablation}d.

\noindent
\textbf{Combination Strategies: } For combination strategies, we evaluated different smoothing filters paired with GFPGAN to analyze their effectiveness in detection evasion and perceptual quality. As shown in Table~\ref{tab:parameter}, all methods achieved high ASRs, with the highest ASR of 76.25\% observed for FabSoften combined with GFPGAN. However, this setting resulted in a slightly lower SSIM (0.8659), reflecting stronger alterations.

\begin{wraptable}{r}{5.5cm}
\caption{Xception AUC (\%) for adversarial train/test configurations on FF++. Bold marks best performance.}\label{tab:xception_adversarial}
\begin{tabular}{lccc}
\toprule
\multirow{4}{*}{\textbf{Train}} & \multicolumn{3}{c}{\textbf{Test}} \\
\cmidrule(lr){2-4}
 & M2 & M4 & M6 \\
\midrule
Pre-trained         & 57.85 & 38.41 & 21.27 \\
M2           & \textbf{89.74} & 45.22 & 24.37 \\
M4 & 69.27 & \textbf{91.85} & 23.36 \\
M6 & 68.19    & 71.38     & \textbf{90.26} \\
\bottomrule
\end{tabular}
\end{wraptable} 

\subsection{Analysis of adversarial training as counter-defense}
We performed adversarial training of the Xception model on the FF++ dataset using different face enhancement methods: Bilateral filter (M2), FabSoften (M4) and FabSoften+GFPGAN (M6). As shown in Table \ref{tab:xception_adversarial}, the pre-trained baseline achieves moderate performance on M2 (57.85\%) but struggled with M4 (38.41\%) and M6 (21.27\%). Fine-tuning on M2 raised detection of M2 substantially to 89.74\% but generalization to other enhancement methods remained limited. In contrast, training with M4 samples achieved 91.85\% on M4 and higher performance on M2 (69.27\%). The model still struggled against M6 achieving only 23.26\% AUC. Finally, training on M6 resulted in more consistent improvements, with 68.19\%, 71.38\%, and 90.26\% across M2, M4 and M6 respectively.

\begin{figure}[t]
    \centering

    \begin{minipage}{0.51\textwidth}
        \centering
        \includegraphics[width=\textwidth]
        {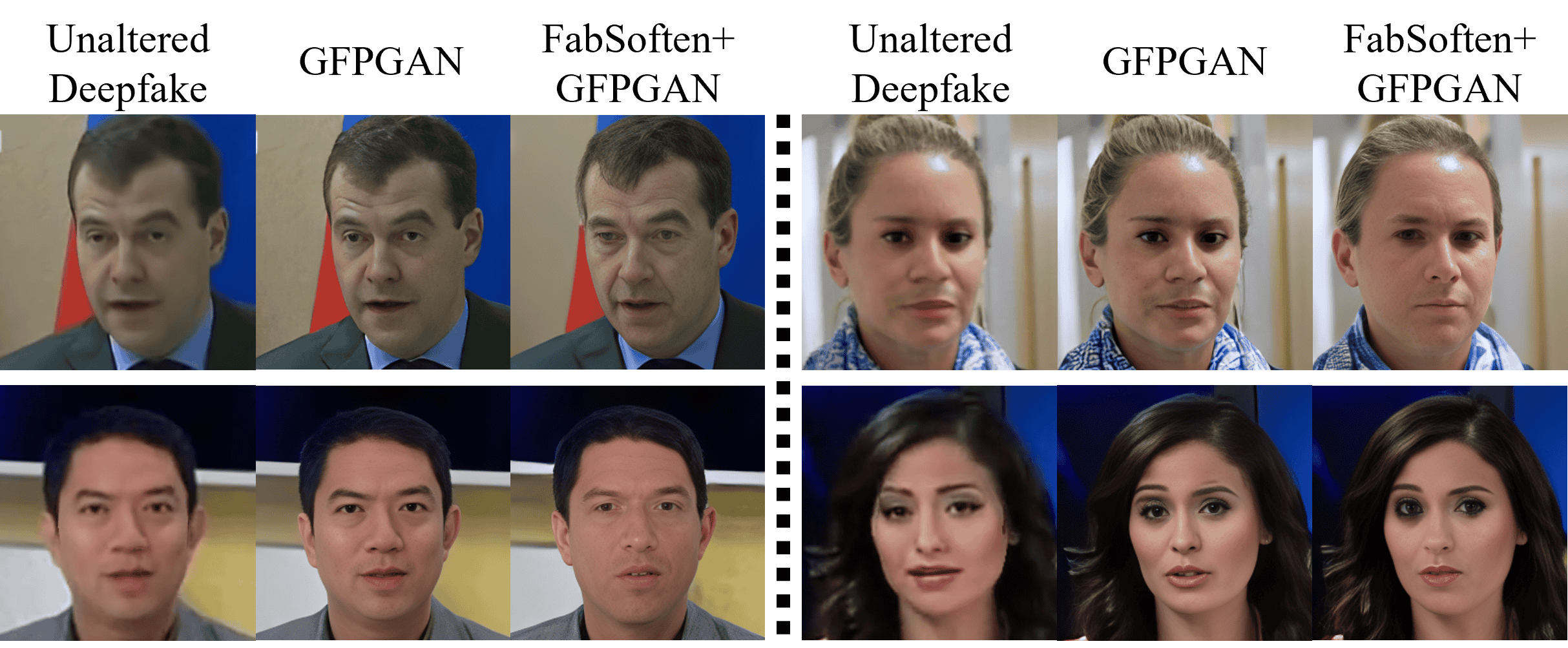}
        {\small (a)}
    \end{minipage}
    \hfill
    \begin{minipage}{0.45\textwidth}
        \centering
        \includegraphics[width=\textwidth]{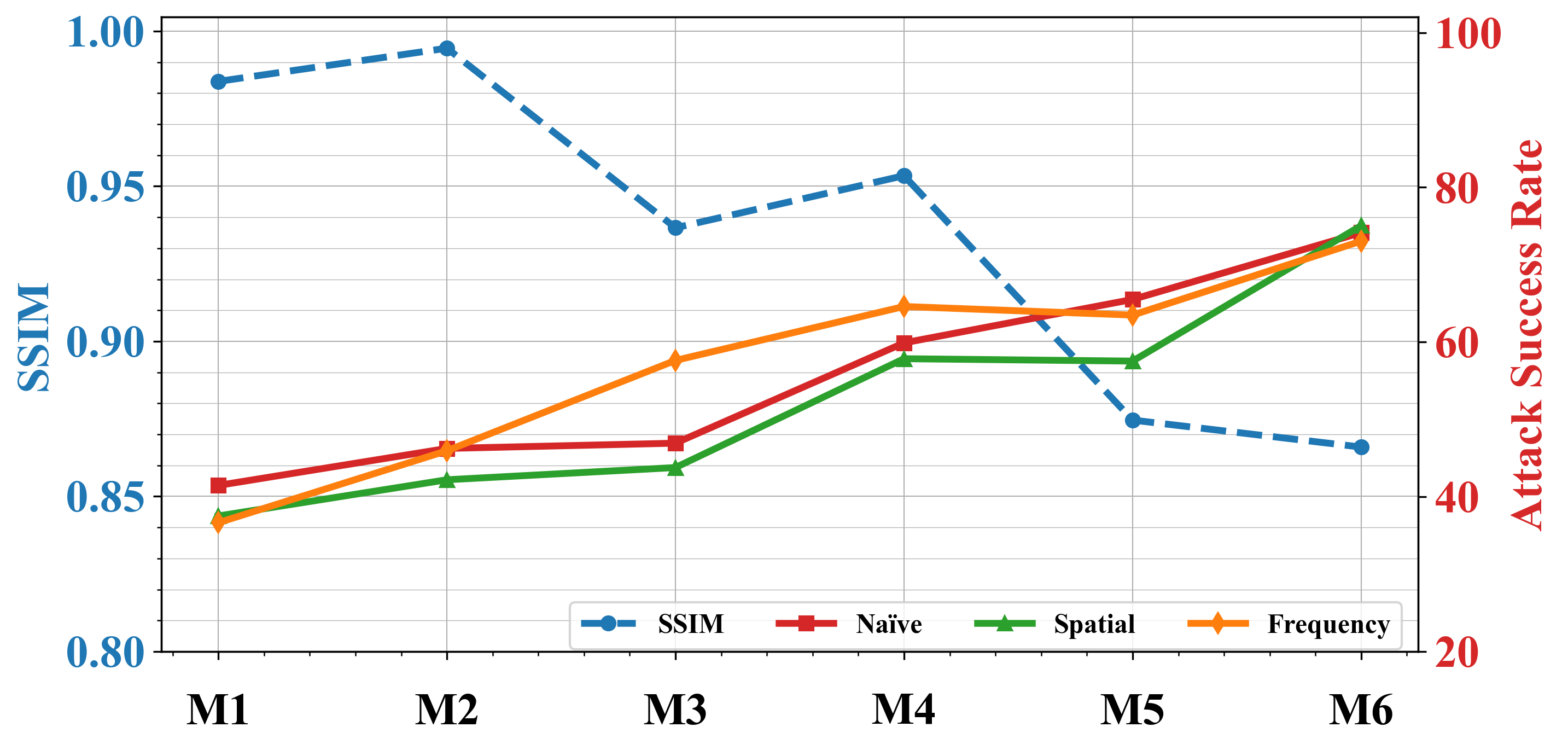}
        {\small (b)}
    \end{minipage}

    \vspace{3mm} 

    \caption{(a) Some severe examples where the GAN-based face enhancement have drastically changed the initial deepfake. Best viewed in color and zoomed in. (b) Trade-off between attack effectiveness and structural similarity across face enhancement methods. }
    \label{fig:trade_severe}
\end{figure}

\section{Discussion}
\noindent
\textbf{Dual Role of face enhancement techniques:}
Our findings underscore the dual nature of face enhancement techniques; while designed to improve visual quality, they inadvertently disrupt biometric cues, thereby misleading deepfake detectors. Designed for tasks such as noise removal, detail restoration, and skin smoothing, these methods enhance image realism and perceptual quality. However, the same transformations that improve appearance can also obscure low-level forensic cues. This suggests the dual-nature of beautification filters in enhancing appearance and hiding synthetic content makes them anti-forensic tools for deepfake defense. As shown in Figure~\ref{fig:trade_severe}b, there exists a consistent trade-off between structural similarity and evasion effectiveness across all detection methods. Even in the absence of malicious intent, these operations can reduce the effectiveness of forensic models by removing the very features on which they rely. \\
\textbf{Ease of use and generalization:}
A notable finding of our study is the ease with which face enhancement techniques can subvert deepfake detection. Unlike adversarial attacks that depend on model access or iterative optimization, these techniques are model-agnostic and operate without any knowledge of the target detector. Most tools are publicly available in consumer apps and editing software, enabling real-world use, especially on social media where content is often enhanced before sharing. Our experiments show that these methods generalize across diverse detectors and datasets, underscoring the need to redesign forensic models for realistic post-processing conditions.


\section{Conclusion}
This study investigated the dual role of face enhancement techniques as both an anti-forensic attack against deepfake detectors and a method to enhance visual realism. Extensive experiments on six deepfake detectors across three benchmark datasets revealed a significant performance degradation. Our evaluation provides a comprehensive analysis of the impact of image processing and GAN-based face enhancement techniques on deepfake detectors across three categories: Naïve, Spatial, and Frequency-based approaches. Furthermore, we evaluated the trade-off between structural similarity and detection evasion, highlighting how these enhancement techniques improve visual realism while simultaneously bypassing detection mechanisms. This study highlight that simple yet effective face enhancement methods can pose new challenges for the forensic models, adding another layer of challenge to the design of deepfake detectors. While adversarial training has shown promising results, it still shows limited generalization. These manipulations introduce distributional shifts rather than adversarial noise, making them more difficult to counter using conventional adversarial defense strategies. In future work, addressing these challenges will enhance  the reliability ad robustness of deepfake detectors in adversarial scenarios.



\begin{thebibliography}{26}
\providecommand{\natexlab}[1]{#1}
\providecommand{\url}[1]{\texttt{#1}}
\expandafter\ifx\csname urlstyle\endcsname\relax
  \providecommand{\doi}[1]{doi: #1}\else
  \providecommand{\doi}{doi: \begingroup \urlstyle{rm}\Url}\fi

\bibitem[Ahmed et~al.(2006)Ahmed, Natarajan, and Rao]{ahmed2006discrete}
Nasir Ahmed, T\_ Natarajan, and Kamisetty~R Rao.
\newblock Discrete cosine transform.
\newblock \emph{IEEE transactions on Computers}, 100:\penalty0 90--93, 2006.

\bibitem[Ain et~al.(2024)Ain, Javed, Malik, and Irtaza]{ain2024exposing}
Qurat~Ul Ain, Ali Javed, Khalid~Mahmood Malik, and Aun Irtaza.
\newblock Exposing the limits of deepfake detection using novel facial mole attack: A perceptual black-box adversarial attack study.
\newblock In \emph{2024 IEEE International Conference on Image Processing (ICIP)}, pages 3820--3826. IEEE, 2024.

\bibitem[Carlini and Farid(2020)]{carlini2020evading}
Nicholas Carlini and Hany Farid.
\newblock Evading deepfake-image detectors with white-and black-box attacks.
\newblock In \emph{Proceedings of the IEEE/CVF conference on computer vision and pattern recognition workshops}, pages 658--659, 2020.

\bibitem[Carlini and Wagner(2017)]{carlini2017towards}
Nicholas Carlini and David Wagner.
\newblock Towards evaluating the robustness of neural networks.
\newblock In \emph{2017 ieee symposium on security and privacy (sp)}, pages 39--57. Ieee, 2017.

\bibitem[Dong et~al.(2023)Dong, Wang, Lai, and Xie]{dong2023restricted}
Junhao Dong, Yuan Wang, Jianhuang Lai, and Xiaohua Xie.
\newblock Restricted black-box adversarial attack against deepfake face swapping.
\newblock \emph{IEEE Transactions on Information Forensics and Security}, 18:\penalty0 2596--2608, 2023.

\bibitem[Gowrisankar and Thing(2024)]{gowrisankar2024adversarial}
Balachandar Gowrisankar and Vrizlynn~LL Thing.
\newblock An adversarial attack approach for explainable ai evaluation on deepfake detection models.
\newblock \emph{Computers \& Security}, 139:\penalty0 103684, 2024.

\bibitem[Hou et~al.(2023)Hou, Guo, Huang, Xie, Ma, and Zhao]{hou2023evading}
Yang Hou, Qing Guo, Yihao Huang, Xiaofei Xie, Lei Ma, and Jianjun Zhao.
\newblock Evading deepfake detectors via adversarial statistical consistency.
\newblock In \emph{Proceedings of the IEEE/CVF Conference on Computer Vision and Pattern Recognition}, pages 12271--12280, 2023.

\bibitem[Huang et~al.(2020)Huang, Juefei-Xu, Wang, Guo, Ma, Xie, Li, Miao, Liu, and Pu]{huang2020fakepolisher}
Yihao Huang, Felix Juefei-Xu, Run Wang, Qing Guo, Lei Ma, Xiaofei Xie, Jianwen Li, Weikai Miao, Yang Liu, and Geguang Pu.
\newblock Fakepolisher: Making deepfakes more detection-evasive by shallow reconstruction.
\newblock In \emph{Proceedings of the 28th ACM international conference on multimedia}, pages 1217--1226, 2020.

\bibitem[Hussain et~al.(2021)Hussain, Neekhara, Jere, Koushanfar, and McAuley]{hussain2021adversarial}
Shehzeen Hussain, Paarth Neekhara, Malhar Jere, Farinaz Koushanfar, and Julian McAuley.
\newblock Adversarial deepfakes: Evaluating vulnerability of deepfake detectors to adversarial examples.
\newblock In \emph{Proceedings of the IEEE/CVF winter conference on applications of computer vision}, pages 3348--3357, 2021.

\bibitem[Ivanovska and Struc(2024)]{ivanovska2024vulnerability}
Marija Ivanovska and Vitomir Struc.
\newblock On the vulnerability of deepfake detectors to attacks generated by denoising diffusion models.
\newblock In \emph{Proceedings of the IEEE/CVF winter conference on applications of computer vision}, pages 1051--1060, 2024.

\bibitem[Kwok and Koh(2021)]{kwok2021deepfake}
Andrei~OJ Kwok and Sharon~GM Koh.
\newblock Deepfake: a social construction of technology perspective.
\newblock \emph{Current Issues in Tourism}, 24\penalty0 (13):\penalty0 1798--1802, 2021.

\bibitem[Li et~al.(2023)Li, Li, Yang, and Deng]{li2023frequency}
Tengjiao Li, Maosen Li, Yanhua Yang, and Cheng Deng.
\newblock Frequency domain regularization for iterative adversarial attacks.
\newblock \emph{Pattern Recognition}, 134:\penalty0 109075, 2023.

\bibitem[Luo et~al.(2021)Luo, Zhang, Yan, and Liu]{luo2021generalizing}
Yuchen Luo, Yong Zhang, Junchi Yan, and Wei Liu.
\newblock Generalizing face forgery detection with high-frequency features.
\newblock In \emph{Proceedings of the IEEE/CVF conference on computer vision and pattern recognition}, pages 16317--16326, 2021.

\bibitem[Meng et~al.(2024)Meng, Wang, Guo, Ju, and Zhao]{meng2024ava}
Xiangtao Meng, Li~Wang, Shanqing Guo, Lei Ju, and Qingchuan Zhao.
\newblock Ava: Inconspicuous attribute variation-based adversarial attack bypassing deepfake detection.
\newblock In \emph{2024 IEEE Symposium on Security and Privacy (SP)}, pages 74--90. IEEE, 2024.

\bibitem[Ni et~al.(2022)Ni, Meng, Yu, Quan, Ren, and Zhao]{ni2022core}
Yunsheng Ni, Depu Meng, Changqian Yu, Chengbin Quan, Dongchun Ren, and Youjian Zhao.
\newblock Core: Consistent representation learning for face forgery detection.
\newblock In \emph{Proceedings of the IEEE/CVF conference on computer vision and pattern recognition}, pages 12--21, 2022.

\bibitem[Qian et~al.(2020)Qian, Yin, Sheng, Chen, and Shao]{qian2020thinking}
Yuyang Qian, Guojun Yin, Lu~Sheng, Zixuan Chen, and Jing Shao.
\newblock Thinking in frequency: Face forgery detection by mining frequency-aware clues.
\newblock In \emph{European conference on computer vision}, pages 86--103. Springer, 2020.

\bibitem[Rossler et~al.(2019)Rossler, Cozzolino, Verdoliva, Riess, Thies, and Nie{\ss}ner]{rossler2019faceforensics++}
Andreas Rossler, Davide Cozzolino, Luisa Verdoliva, Christian Riess, Justus Thies, and Matthias Nie{\ss}ner.
\newblock Faceforensics++: Learning to detect manipulated facial images.
\newblock In \emph{Proceedings of the IEEE/CVF international conference on computer vision}, pages 1--11, 2019.

\bibitem[Sha et~al.(2023)Sha, Zhang, Shen, Li, and Mei]{sha2023deep}
Tong Sha, Wei Zhang, Tong Shen, Zhoujun Li, and Tao Mei.
\newblock Deep person generation: A survey from the perspective of face, pose, and cloth synthesis.
\newblock \emph{ACM Computing Surveys}, 55\penalty0 (12):\penalty0 1--37, 2023.

\bibitem[Tan and Le(2019)]{tan2019efficientnet}
Mingxing Tan and Quoc Le.
\newblock Efficientnet: Rethinking model scaling for convolutional neural networks.
\newblock In \emph{International conference on machine learning}, pages 6105--6114. PMLR, 2019.

\bibitem[Taubin(1995)]{taubin1995curve}
Gabriel Taubin.
\newblock Curve and surface smoothing without shrinkage.
\newblock In \emph{Proceedings of IEEE international conference on computer vision}, pages 852--857. IEEE, 1995.

\bibitem[Tomasi and Manduchi(1998)]{tomasi1998bilateral}
Carlo Tomasi and Roberto Manduchi.
\newblock Bilateral filtering for gray and color images.
\newblock In \emph{Sixth international conference on computer vision (IEEE Cat. No. 98CH36271)}, pages 839--846. IEEE, 1998.

\bibitem[Uddin et~al.(2023{\natexlab{a}})Uddin, Yang, Jeong, and Oh]{uddin2023robust}
Kutub Uddin, Yoonmo Yang, Tae~Hyun Jeong, and Byung~Tae Oh.
\newblock A robust open-set multi-instance learning for defending adversarial attacks in digital image.
\newblock \emph{IEEE Transactions on Information Forensics and Security}, 2023{\natexlab{a}}.

\bibitem[Uddin et~al.(2023{\natexlab{b}})Uddin, Yang, and Oh]{uddin2023deep}
Kutub Uddin, Yoonmo Yang, and Byung~Tae Oh.
\newblock Deep learning-based counter anti-forensic of gan-based attack in hevc compressed domain using coding pattern analysis.
\newblock \emph{Expert Systems with Applications}, 233:\penalty0 120912, 2023{\natexlab{b}}.

\bibitem[Velusamy et~al.(2020)Velusamy, Parihar, Kini, and Rege]{velusamy2020fabsoften}
Sudha Velusamy, Rishubh Parihar, Raviprasad Kini, and Aniket Rege.
\newblock Fabsoften: Face beautification via dynamic skin smoothing, guided feathering, and texture restoration.
\newblock In \emph{Proceedings of the IEEE/CVF Conference on Computer Vision and Pattern Recognition Workshops}, pages 530--531, 2020.

\bibitem[Wang et~al.(2021)Wang, Li, Zhang, and Shan]{wang2021towards}
Xintao Wang, Yu~Li, Honglun Zhang, and Ying Shan.
\newblock Towards real-world blind face restoration with generative facial prior.
\newblock In \emph{Proceedings of the IEEE/CVF conference on computer vision and pattern recognition}, pages 9168--9178, 2021.

\bibitem[Yan et~al.(2023)Yan, Zhang, Fan, and Wu]{yan2023ucf}
Zhiyuan Yan, Yong Zhang, Yanbo Fan, and Baoyuan Wu.
\newblock Ucf: Uncovering common features for generalizable deepfake detection.
\newblock In \emph{Proceedings of the IEEE/CVF International Conference on Computer Vision}, pages 22412--22423, 2023.

\end{thebibliography}
\end{document}